\documentclass[letterpaper, 10 pt, conference]{ieeeconf}  %

\IEEEoverridecommandlockouts                              %
\makeatletter
\let\NAT@parse\undefined
\makeatother

\usepackage{url}

\usepackage{makeidx}         %
\usepackage{graphicx}        %
\usepackage{multicol}        %
\usepackage{amsmath}
\usepackage{amssymb}
\usepackage{booktabs}
\usepackage{xspace}
\usepackage{color}
\usepackage[hang,flushmargin]{footmisc}
\usepackage{microtype}
\usepackage{float}
\usepackage{array}
\usepackage{multirow}
\usepackage{flushend}
\usepackage{cite}

\usepackage{cite}
\usepackage{amsmath,amssymb,amsfonts}
\usepackage{algorithmic}
\usepackage{graphicx}
\usepackage{textcomp}
\usepackage{xcolor}

\usepackage{booktabs}
\usepackage{outlines}
\usepackage{caption}
\usepackage{subcaption}
\usepackage[hang,flushmargin]{footmisc}
\usepackage{stmaryrd}

\usepackage{microtype}
\usepackage{hhline}
\usepackage{multirow}
\usepackage{siunitx}
\usepackage{makecell}
\usepackage{gensymb}
\usepackage{placeins}

\definecolor{cvprblue}{rgb}{0.21,0.49,0.74}
\usepackage[breaklinks,colorlinks]{hyperref}

\usepackage{tensor}

\DeclareMathAlphabet{\mathcal}{OMS}{cmsy}{m}{n}

\renewcommand{\vec}[1]{\ensuremath{\mathbf{#1}}}
\newcommand{\mat}[1]{\ensuremath{\mathbf{#1}}}
\newcommand{\norm}[1]{\left\lVert#1\right\rVert}

\newcommand{\prob}[1]{\ensuremath{p\left(#1\right)}}
\newcommand{\probnobr}[1]{\ensuremath{p(#1)}}           %
\newcommand{\probc}[2]{\ensuremath{\prob{#1 \;\middle\vert\; #2}}}

\newcommand{\set}[1]{\ensuremath{\left\{#1\right\}}}

\newcommand{\tf}[3]{\tensor[^{#1}]{\mat{#2}}{_{#3}}}

\usepackage{xspace}

\renewcommand{\[}{\begin{equation}}
\renewcommand{\]}{\end{equation}}

\usepackage[capitalize]{cleveref}
\crefname{section}{Sec.}{Secs.}
\Crefname{section}{Section}{Sections}
\Crefname{table}{Table}{Tables}
\crefname{table}{Tab.}{Tabs.}
\crefname{algorithm}{Algo.}{Algos.}
\Crefname{algorithm}{Algorithm}{Algorithms}

\definecolor{red}{RGB}{255, 0, 0}   %
\definecolor{orange}{RGB}{255, 77, 0}   %
\definecolor{green}{RGB}{0, 128, 0}   %
\definecolor{purple}{RGB}{160, 32, 240}   %
\definecolor{lightblue}{RGB}{52, 155, 235}   %
\definecolor{darkmagenta}{RGB}{204, 51, 139} %

\def\BibTeX{{\rm B\kern-.05em{\sc i\kern-.025em b}\kern-.08em
    T\kern-.1667em\lower.7ex\hbox{E}\kern-.125emX}}

\newcommand{\ourName}{PseudoTouch}

\newcommand{\ourNameMath}{PT}

\newcommand{\realSpace}{\mathbb{R}}

\newcommand{\reskin}{ReSkin\xspace}

\newcommand{\allObjects}{O}
\newcommand{\singleObject}{o}

\newcommand{\mleft}{l}
\newcommand{\mright}{r}

\newcommand{\pointCloud}{\mathcal{P}}
\newcommand{\point}{p}

\newcommand{\allTouches}{N}
\newcommand{\touchIndex}{n}

\newcommand{\depthPatch}{d}

\newcommand{\touchLocation}{l}
\newcommand{\touchLocationPredicted}{\tilde{\touchLocation}}
\newcommand{\touchLocationProbability}{\alpha}
\newcommand{\normal}{n}

\newcommand{\touchResult}{\tau}
\newcommand{\touchPredicted}{\tilde{\touchResult}}

\newcommand{\graspQualityNetwork}{GQ}
\newcommand{\graspQualityNetworkTouch}{\graspQualityNetwork_\touchResult}
\newcommand{\graspQualityNetworkPC}{\graspQualityNetwork_\pointCloud}
\newcommand{\graspQualityOutput}{s}

\usepackage{pifont}%

\def\ourmodel{PseudoTouch\xspace}

\title{\LARGE \bf
    \ourmodel: Efficiently Imaging the Surface Feel of Objects for\\Robotic Manipulation
}

\author{Adrian Röfer$^{*1}$, Nick Heppert$^{*1,2}$, Abdallah Ayad$^{1}$, Eugenio Chisari$^{1}$, Abhinav Valada$^{1}$
\thanks{$^*$ These authors contributed equally.}
\thanks{
    \mbox{$^1$}University of Freiburg, Germany, 
    \mbox{$^2$Zuse} School ELIZA
}%
\thanks{This work was funded by the Carl Zeiss Foundation with the ReScaLe project and the BrainLinks-BrainTools center of the University of Freiburg. Nick Heppert is supported by the Konrad Zuse School of Excellence in Learning and Intelligent Systems (ELIZA) through the DAAD programme Konrad Zuse Schools of Excellence in Artificial Intelligence, sponsored by the Federal Ministry of Education and Research.}
}
    
\begin{document}

\maketitle

\begin{abstract}
Tactile sensing is vital for human dexterous manipulation, however, it has not been widely used in robotics. Compact, low-cost sensing platforms can facilitate a change, but unlike their popular optical counterparts, they are difficult to deploy in high-fidelity tasks due to their low signal dimensionality and lack of a simulation model. To overcome these challenges, we introduce \ourmodel which links high-dimensional structural information to low-dimensional sensor signals. It does so by learning a low-dimensional visual-tactile embedding, wherein we encode a depth patch from which we decode the tactile signal. We collect and train \ourmodel on a dataset comprising aligned tactile and visual data pairs obtained through random touching of eight basic geometric shapes. We demonstrate the utility of our trained \ourmodel model in two downstream tasks: object recognition and grasp stability prediction. In the object recognition task, we evaluate the learned embedding's performance on a set of five basic geometric shapes and five household objects. Using \ourmodel, we achieve an object recognition accuracy $84\%$ after just ten touches, surpassing a proprioception baseline. For the grasp stability task, we use ACRONYM labels to train and evaluate a grasp success predictor using \ourmodel's predictions derived from virtual depth information. Our approach yields a $32\%$ absolute improvement in accuracy compared to the baseline relying on partial point cloud data. We make the data, code, and trained models publicly available at \url{https://pseudotouch.cs.uni-freiburg.de}.
\end{abstract}

\section{Introduction}
\label{sec:intro}

Successful dexterous object manipulation requires the ability to identify an object and track its state throughout the task. Robust vision-based methods for object detection~\cite{lang2022robust}, pose estimation~\cite{chisari2023centergrasp}, and tracking~\cite{von2023treachery} have become ubiquitous in robotics. However, vision is an undependable modality: the view of the object can be partially or fully occluded during the motion, or the space from which the object should be grasped can be completely unobservable. Moreover, the visual resolution can be insufficient for tracking a small object. 
In contrast when Humans are faced with such situations, they can proceed due to their refined and integrated visual-tactile sensing abilities~\cite{park2018recent}. Thus, unlocking similar capabilities for robotics is key for general purpose manipulators.

\begin{figure}
    \centering
    \includegraphics[width=\columnwidth]{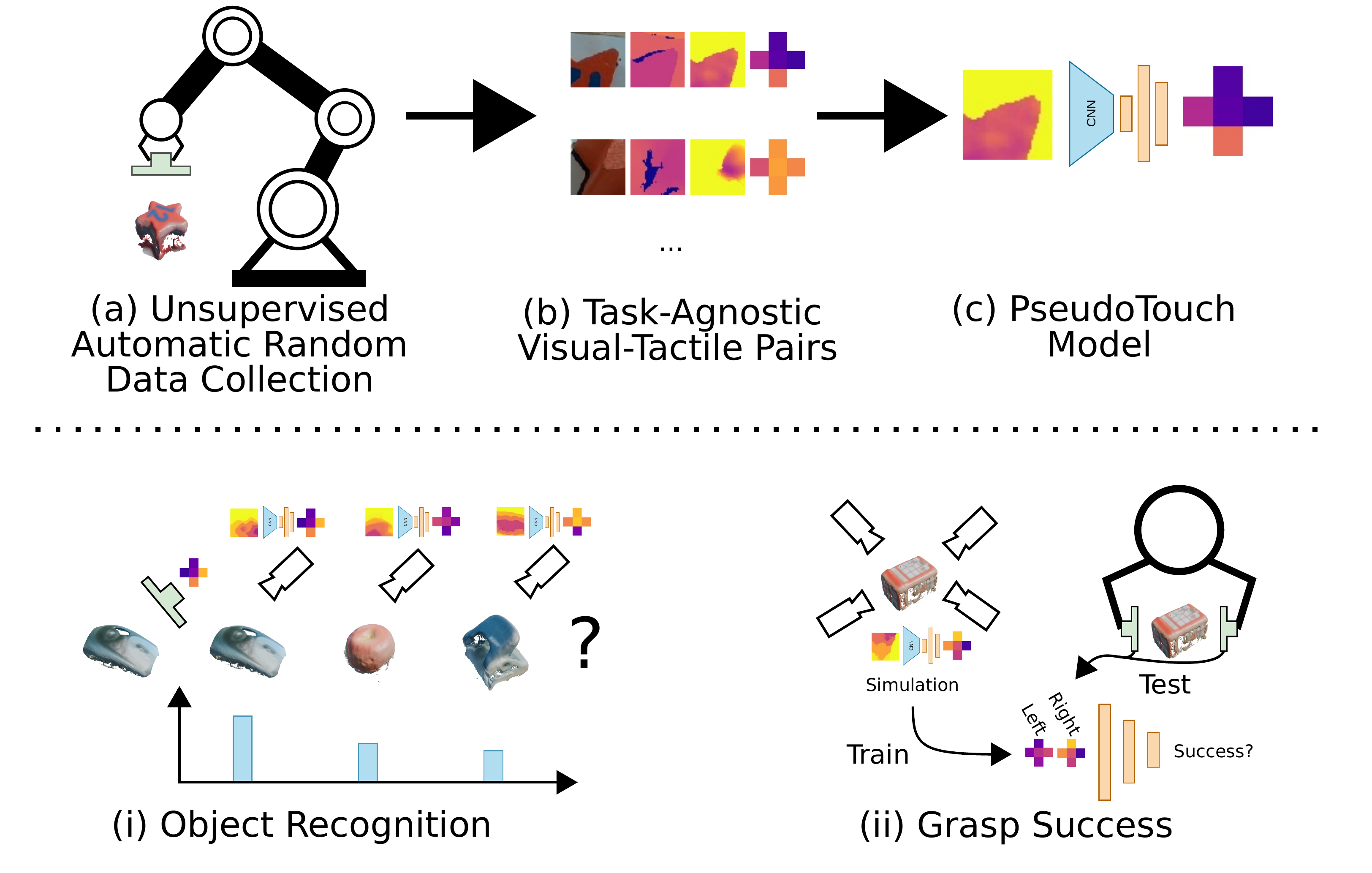}
    \caption{Overview of the \ourName{} model~which infers the tactile signal given a visual input image. We use an unsupervised automatic data collection~\textit{(a)} to generate a dataset of task-agnostic visual-tactile pairs~\textit{(b)} used for training the model~\textit{(c)}. We demonstrate the model's utility in two downstream tasks. First, for object recognition~\textit{(i)} in which we compare \ourName's generated touch readings of potential objects to measured readings. Second, we use \ourName{} to automatically enhance grasp labels with touch readings in a grasp stability prediction task~\textit{(ii)}.}
    \label{fig:teaser}
    \vspace{-3mm}
\end{figure}

Nonetheless, while tactile sensing is not new to robotics, learning-based efforts have given much attention to optical tactile sensors. Here, existing works have used tactile embeddings for in-hand object pose estimation~\cite{bauza2022tac2pose, buchanan2023online}, object recognition~\cite{xu2023tandem3d}, and for improving robotic grasping~\cite{chumbley_integrating_2022}. However, the aforementioned works rely on tactile perception through optical sensing systems, which have the disadvantages of high cost and a bulky form factor. Furthermore, the high dimensionality of the optical sensing scheme requires a large number of training samples. On the other hand, there has recently been a surge in small, low-cost sensor designs that produce low-dimensional but high-frequency measurements of magnetic fields~\cite{bhirangi2021reskin,yan2021soft}. These designs are attractive, as they seem more accessible and easier to integrate into robotic systems. However, they have still not found wide adoption in robotics, likely due to two main challenges they present. \textit{First}, unlike their optical counterparts, these sensors are challenging to simulate. While optical sensors can be imitated using standard rendering techniques, no such standard solution exists for the magnetic fields generated by these sensors. Works such as~\cite{bauza2022tac2pose} rely on simulated examples to train their models efficiently. \textit{Second}, the sensors' low signal dimensionality casts doubt on their utility for high-fidelity tasks requiring distinguishing fine surface details. In~\cite{bhirangi2021reskin,yan2021soft}, these sensor designs estimate a contact point and force, but detailed structural sensing as in~\cite{bauza2022tac2pose} remains largely underexplored with Imagine2Touch~\cite{ayad2024} being one of the first works showing the potential of the low signals dimensionality.\looseness=-1

In this work, we address these two challenges by proposing, \ourName, a sample-efficient approach for connecting high-dimensional structural information with low-dimensional tactile information through a learned function. Our goal is to provide a sample-efficient method that, once trained, can be used across multiple robotic tasks. Specifically, we use the \reskin sensor~\cite{bhirangi2021reskin} as a platform for which we introduce a data-collection and simulation-augmented training scheme for predicting touch readings from depth images. We train \ourName{} on a few elementary geometric features as depicted in \cref{fig:teaser} and evaluate the utility on two downstream tasks: \textit{object recognition} and \textit{grasp stability prediction}. We show that despite the low-dimensional sensor readings, \ourName{} enables competitive results on these tasks. We additionally conduct a user study to compare our model to a human touch baseline, whose result highlights the difficulty of the tasks and the effectiveness of our approach.

The main contributions are as follows:
\begin{outline}
    \1 A model architecture for successfully inferring tactile readings from depth images.
    \1 A data-collection and training setup that allows training our model using a small amount of data.
    \1 Two demonstrator applications were used to evaluate the utility of \ourName: an object recognition task and a grasp quality prediction task.
    \1 Code, trained models, and data are made publicly available at \url{https://pseudotouch.cs.uni-freiburg.de}.
\end{outline}

\section{Related Work}
\label{sec:related_work}
This section briefly reviews previous visual-tactile cross-modality learning approaches as well as introduces current approaches for our proposed downstream tasks, and shows how \ourmodel can enhance these approaches.

\subsection{Visual-Tactile Embeddings}
\label{sec:related_work:visual_tactile_embeddings}
Combining different modalities through a common embedding has been widely researched~\cite{younes2023catch, hurtado2022semantic}. Specifically, the combination of visual and tactile modalities has been used for reconstructing 3D shapes~\cite{rustler_active_2022, smith_3d_2020}. Here, typically, the choice for fusing both modalities is at the point cloud level~\cite{murali_touch_2023}. While \cite{falco_cross-modal_2017} still uses point clouds as their fusion representation, they investigate how to define descriptors for both the visual point clouds and tactile point clouds. In a follow-up work~\cite{falco2019transfer}, the authors investigate how to minimize a distribution shift between both descriptors.
\cite{lee_making_2020} fuses both modalities in a common embedding space trained end-to-end and used for a downstream robotic manipulation task, whereas \cite{zhong_touching_2023} learns an implicit representation of shape and tactile through a neural radiance field~\cite{mildenhall2021nerf}, which first renders color and depth images given a camera pose and then generate the corresponding touch reading.
Closest to our work is \cite{lee_touching_2019, yang_generating_2023}, which predicts the touch signal from an RGB image and vice versa. \cite{yang_binding_2024} go even one step further and include audio and text.
Opposed to other works, the \reskin sensor we use is low dimensional and cannot be easily transformed into a local point cloud, making previous object-centric representations unsuitable. Thus, we learn a cross-modal embedding focused on a small local patch, which can be used, as demonstrated in our object recognition task, to build an object-centric representation again.

\subsection{Down Stream Tasks}
\label{sec:related_work:down_stream_tasks}
There have been several uses of learned tactile models in robotics.
For a common robotics task, \cite{bauza2022tac2pose} performs in-hand object recognition and pose estimation by training an embedding model on simulated data using CAD models. During inference, they infer the correct object pose distribution by matching their embedding of the sensor data with the embeddings of their simulated data.
Using the \reskin sensor, \cite{bhirangi2022all} build a low-cost robotic hand with tactile sensing. Despite the low-dimensional sensor signal, they are able to train classifiers for material, texture, and softness identification. They generate data using a simple motor-babbling policy and feed this data through an LSTM followed by multiple linear layers with ReLU activations. They achieve $65\%$ accuracy in these tasks when deploying their models on unseen objects.

{\parskip=2pt\noindent \textit{Object Recognition}:}
Most closely related to our downstream task of object recognition, \cite{xu2023tandem3d} addresses object recognition using tactile sensing. Using a sensor that can only measure a contact point and the surface normal, they train an exploration policy and a discriminator to propose new sample points and to predict the object class. Both of these modules work off of a point cloud encoder, which encodes the already sampled contact points and their normals. They show very high accuracy in object identification while requiring significantly fewer sample points than their baselines.
Also similar to our downstream task of object recognition, \cite{corcodel_interactive_2020} study the problem of learning object recognition using an array of pressure sensors. They focus on building an incremental tactile model of objects by having the robot explore them and store the information in a 3D representation. This differs from our downstream task in that we are proposing a pretrained model that can predict these readings outright from uniformed touches.\looseness=-1

{\parskip=2pt\noindent \textit{Grasp Stability Prediction}:}
Previously \cite{chumbley_integrating_2022, calandra_feeling_2017} showed that predicting grasp stability with tactile signals outperforms a pure visual baseline. While \cite{chumbley_integrating_2022} were able to collect a large dataset in simulation as they used a simulatable vision-based tactile sensor, \cite{calandra_feeling_2017} collected a large real-world dataset. 
By using \ourmodel to generate tactile readings for simulated visual patches, we can use pre-annotated grasp stability datasets to generate a dataset, avoiding the need to either simulate a touch sensor or collect extensive real-world grasp labels with a tactile sensor.

\section{Technical Approach}
\label{sec:technical_approach}
We first detail our \ourName{} approach and then describe how we use its predictions in the two downstream tasks we use to evaluate its utility.

\subsection{\ourName{}-Model}
\label{sec:technical_approach:our_model}
Our proposed model is a function $\ourNameMath : z_d \Rightarrow \tilde{\tau}$, which takes a depth image $z_d \in \realSpace^{17\times17}$ of the object surface as input and predicts the tactile reading that would be emitted touching the surface $\tilde{\tau} \in \realSpace^{15}$.
We implement this function as a neural network, consisting of a light-weight convolutional encoder, followed by a flattening average pooling, the output of which is fed through a single layer MLP before predicting the tactile reading $\tilde{\tau}$.
We chose a very low-capacity model to prevent overfitting on our small dataset due to the large size imbalance between the input and output of our network.\looseness=-1

\begin{figure}
    \vspace{2mm}
    \centering
    \includegraphics[width=0.4\linewidth]{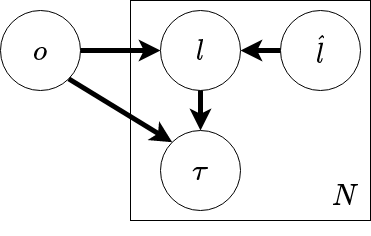}
    \caption{Object Recognition Model. $\singleObject$ is an object hypothesis, $\touchLocation$ is the location where the actual touch was recorded, $\touchLocationPredicted$ is the location where we intended to touch and $\touchResult$ is the actual result of our touch. $\allTouches$ is the number of all touches performed. 
    }
    \label{fig:object_recognition_model}
    \vspace{-3mm}
\end{figure} 

\subsection{Online Object Recognition}
\label{sec:technical_approach:object_recognition}
In our first task, we use \ourmodel to recognize objects from a given set of candidate objects by subsequently performing $\allTouches$ touches and updating our current object hypothesis based on the touch result through a graphical model.\looseness=-1 %

{\noindent \textbf{Algorithmic Outline}}:
We assume we are given a set of objects $\allObjects$ with known 3D representations (e.g. meshes), denoted by $\pointCloud_\singleObject$ for object $\singleObject$. Up to our maximum number of touches $\allTouches$, we calculate the current object likelihood given the current observations consisting of previous touch \mbox{locations $L = \set{l_1, \ldots, l_i}$} and \mbox{readings $T = \set{\tau_1, \ldots, \tau_i}$}. We sample from that distribution to get an object hypothesis $\tilde{o}$ on which we will randomly sample a location $\touchLocationPredicted$ with an associated surface normal~$\normal$. 
To perform the touch at the specified location, we move the robot there along the normal~$\normal$ until the measured force exceeds a threshold. We then record the position of the end-effector $l_{i+1}$ and touch signal $\tau_{i+1}$ and add them to the pool of observations.

{\noindent \textbf{Graphical Model Formulation}}: We formulate the object recognition task as a classical directed graphical model (see \cref{fig:object_recognition_model}). 
The full probability for a single touch is given through 
\begin{align}
    \label{eqn:}
    \prob{\singleObject, \touchLocationPredicted, \touchLocation, \touchResult} = 
    \prob{\singleObject}
    \probc{\touchResult}{\touchLocation, \singleObject}
    \probc{\touchLocation}{\singleObject}
    \probc{\touchLocation}{\touchLocationPredicted}
    \prob{\touchLocationPredicted}
\end{align}
assuming conditional independence between the object $\singleObject$ and $\touchLocationPredicted$. In the following, we define each probability term, starting from left to right.

Initially, we assume a uniform prior for each object $\singleObject$
\begin{equation}
    \label{eqn:prior}
    \prob{\singleObject} = \frac{1}{|\allObjects|}
\end{equation}
over the set of all objects $O$. Next, we introduce probabilities for touch and proprioceptive observations. 
Given the reading of a touch $\touchResult$, we formulate the touch likelihood for each object $\singleObject$ as
\begin{equation}
    \probc{\touchResult}{\singleObject, \touchLocation} = \exp\left(- \norm{\touchPredicted_o - \touchResult}\right),
\end{equation}
where \ourmodel is used to predict a hypothesized touch $\touchPredicted_\singleObject = \ourNameMath \left( \tilde{z}_d(\singleObject, \touchLocation) \right)$ for an object $\singleObject$ at location $\touchLocation$ based on the simulated depth patch $\tilde{z}_d(\singleObject, \touchLocation)$.
Similarly, we define the proprioceptive likelihood $\probc{\touchLocation}{\singleObject}$ as
\begin{equation}
    \probc{\touchLocation}{\singleObject} = \exp\left( - \norm{\min( \touchLocation, \pointCloud_\singleObject)} \right),
\end{equation}
where $\min( \touchLocation, \pointCloud) = \min_{\point \in \pointCloud} \norm{\touchLocation - \point}$ is the minimum distance from the location $\touchLocation$ to a point $\point$ on the objects surface $\pointCloud$.
We assume both the probability \mbox{$\probnobr{\touchLocationPredicted} = \touchLocationProbability$} of any desired touch location $\touchLocationPredicted$, as well as the conditional probability \mbox{$\probnobr{\touchLocation | \touchLocationPredicted} = \beta$} of actual location $\touchLocation$ to be constant.
After performing $\allTouches$ touches we will have a set of proprioceptive and touch reading tuples, each given as $(\touchResult_\touchIndex, \touchLocation_\touchIndex)$ for the {$\touchIndex$-th} touch as well as desired touch location $(\touchLocationPredicted_\touchIndex)$, which we group as $x_\touchIndex = (\touchResult_\touchIndex, \touchLocation_\touchIndex, \touchLocationPredicted_\touchIndex)$. Assuming we performed $\allTouches$ touches in total, we formulate the full joint probability of an object $\singleObject$ as
\begin{align}
\label{eqn:full_probability}
\begin{split}
    \prob{
        \singleObject,
        x_1, \ldots, x_\allTouches, 
    }
    =
    \prob{\singleObject} 
    \prod_{\touchIndex} 
        \probc{\touchResult_\touchIndex}{\touchLocation_\touchIndex, \singleObject}
        \probc{\touchLocation_\touchIndex}{\singleObject}
\end{split}
\end{align}
already leaving out $\probnobr{ \touchLocation_\touchIndex | \touchLocationPredicted_\touchIndex }$ and $\probnobr{\touchLocationPredicted_\touchIndex}$ for clarity as they will be eventually eliminated in a normalization step. Further, the probability of an object $\singleObject$ given the current observations is described by
\begin{align}
\begin{split}
    \probc{\singleObject}{
        x_1, \ldots, x_\allTouches
    } 
    = 
    \frac{
        \prob{
            \singleObject,
            x_1, \ldots, x_\allTouches
        }
    }{
        \prob{
            x_1, \ldots, x_\allTouches
        }        
    },
\end{split}
\end{align}
where the denominator is given through marginalizing \cref{eqn:full_probability} over all $\tilde{\singleObject} \in \allObjects$.
In each step we then pick the object
\begin{align}
\begin{split}
    \singleObject = \underset{\tilde{\singleObject}}{\arg\,\max}\,
    \probc{\tilde{\singleObject}}{
        (\touchResult_1, \touchLocation_1), \ldots, (\touchResult_\allTouches, \touchLocation_\allTouches), 
        \touchLocationPredicted_1, \ldots \touchLocationPredicted_\allTouches
    }
\end{split}
\end{align}
that best explains the current observations.

\subsection{Grasp Stability Prediction}
\label{sec:technical_approach:grasp_stability_prediction}
In the second downstream task, we use \ourName{} to predict the grasp success on an object. We setup a neural network $\graspQualityNetworkTouch (\touchResult_{\mleft}, \touchResult_{\mright}) = \graspQualityOutput$ that uses the tactile readings from the left $\touchResult_{\mleft}$ and the right finger $\touchResult_{\mright}$ as input and infers a binary score $\graspQualityOutput \in \left[ 0, 1 \right]$ indicating the potential success of the grasp. We use a small fully connected architecture with two hidden layers with $100$ neurons each. %
To train such a network, the straightforward but time-intensive approach would be to collect a training dataset by executing many grasps on real objects recording the touch signal just before the grasp happens, and labeling them with the observed grasp outcome \cite{calandra_feeling_2017}. In contrast, we leverage our proposed \ourName{} and recent advancements in large-scale simulations to generate a training, validation, and test set for the aforementioned grasp stability predictor solely in simulation. Given an object from the ACRONYM~\cite{eppner_acronym_2021} dataset, we set up a virtual renderer in which we use two orthographic virtual cameras at the tips of the gripper at the grasp $g \in SE(3)$ to render depth patches $z_{d,\mleft}, z_{d,\mright}$ for the left and the right finger. We then use \ourName{} (trained on random touches) to infer a virtual tactile signal. Hence, we generate a large-scale dataset to train and test the grasp stability predictor $\graspQualityNetworkTouch$ without the need to execute and annotate real grasps or explicitly simulate a tactile sensor.\looseness=-1

\section{Data Collection}

\begin{figure}
    \centering
    \includegraphics[width=0.47\columnwidth]{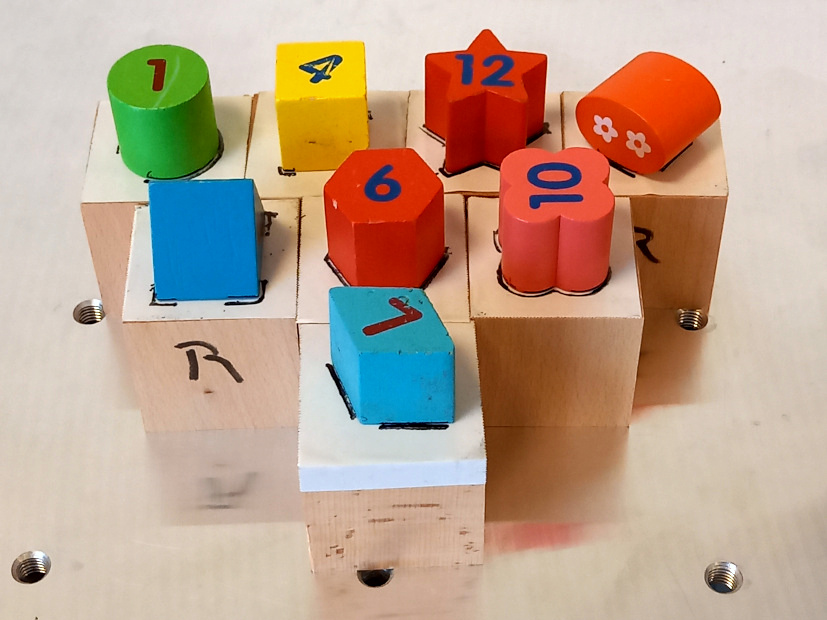}
    \includegraphics[width=0.47\columnwidth]{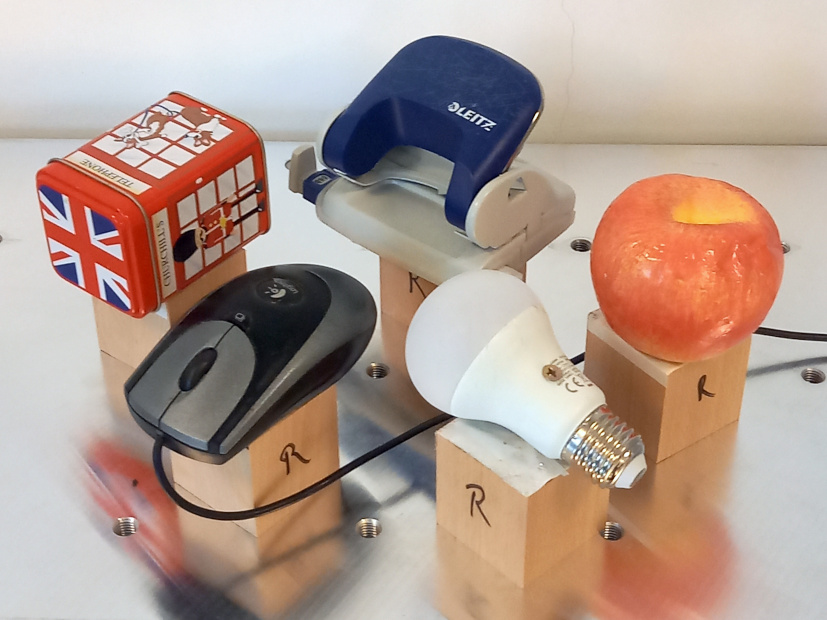}
    \caption{Objects used for training and validation of \ourmodel. Left: Primitive shape objects used for training. Right: Everyday objects are used for object recognition experiments for validation. In clock-wise order starting at the 12 o'clock position: \emph{puncher}, \emph{apple}, \emph{bulb}, \emph{pc mouse}, \emph{tin box}.}
    \label{fig:pt_objects}
    \vspace*{-2mm}
\end{figure}

\begin{figure*}
    \centering
    \begin{subfigure}{0.2\textwidth}
        \centering
        \includegraphics[height=4.5cm]{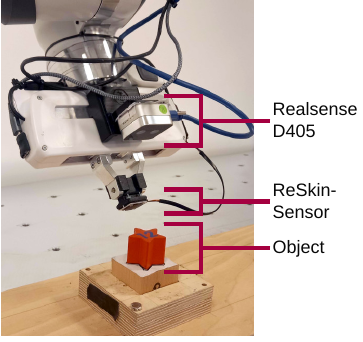}
        \caption{Robotic setup for data collection.}
    \end{subfigure}
    \begin{subfigure}{0.7\textwidth}
        \centering
        \includegraphics[height=4.5cm]{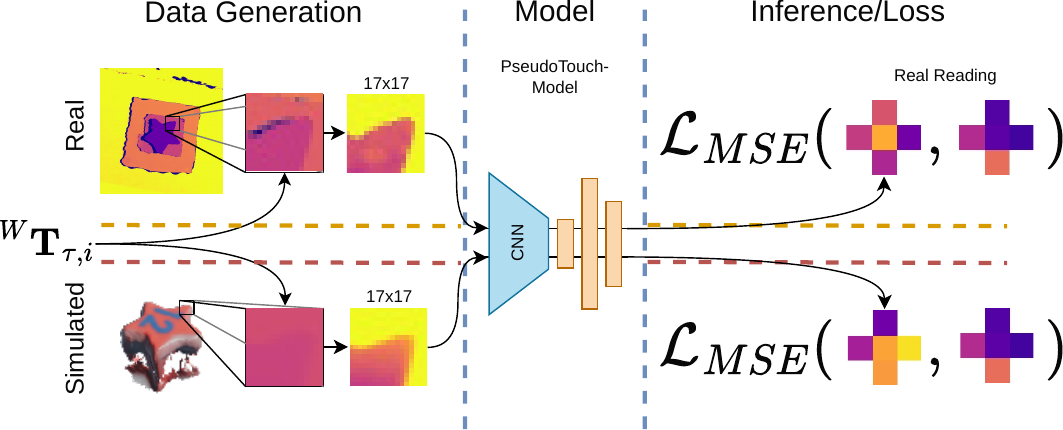}
        \caption{Flow of data from depth image to prediction.}
    \end{subfigure}
    \vspace*{-1mm}
    \caption{\emph{(a)}: Overview of physical robot setup for data collection. The \reskin sensor is attached as a fingertip to the robot's gripper. Using the Realsense~D405 camera, we capture an image before touching the object. The object is attached to a wooden anchor mounted to the table for repeatability. \emph{(b)}: Illustration of data processing and inference. From a real depth image, we crop the section that the robot touched using the recorded end-effector pose $\tf{W}{T}{\tau,i}$. To mitigate the gap between real and simulated data, we use the same pose and a mesh of the object to render a simulated sample. We normalize both depth patches and pass them through our \ourmodel{} model. Finally, we minimize the MSE-loss deviation from the actual sensor reading.}
    \label{fig:setup_and_flow}
    \vspace{-5mm}
\end{figure*}

\label{sec:data_collection}
To train and evaluate \ourmodel, we collect a dataset consisting of tactile readings $\tau \in \realSpace^{15}$ and the depth image $z_d \in \realSpace^{17\times17}$ of the corresponding $\SI{17}{\milli\meter}\times\SI{17}{\milli\meter}$ area that was touched by the sensor. The collection process is semi-automated, as we detail in the following section.

\subsection{Robotic Setup}
\label{sec:data_collection:robotic_setup}
We use a table-mounted Franka Emika robot with a wrist-mounted D405 Realsense RGBD camera and a 3D-printed \emph{finger} holding the \reskin sensor at its end. The objects are fixed on wooden bases which sit firmly in a wooden socket mounted to the table. The whole setup is displayed in \cref{fig:setup_and_flow}.
Our data collection procedure consists of 6 steps:
\begin{enumerate}
    \item Starting with a robot with correct hand-camera calibration, we use Franka's hand-guiding feature to guide the camera around an object and collect $~10$ depth images. We convert these into point clouds, which we prune for all points outside of the object's volume. We integrate the remaining points into a volume~\cite{zhou2013dense}, from which we finally extract a surface mesh $m$.
    \item We sample candidate points $C_m = \set{(\vec{p}_1, \vec{n}_1), \ldots, (\vec{p}_I, \vec{n}_I)}$ from the mesh $m$ by uniformly sampling from it vertices $\vec{p}_i$ and their corresponding normals $\vec{n}_i$. We reject samples if the normal's angle to the table is flatter than $33^\circ$, or we find that the approach corridor along the normal is obstructed.\looseness=-1
    \item To observe the target zone before touching, we define an IK-problem that places the wrist camera along the normal $\vec{n}_i$, $\SI{12}{\centi\meter}$ away from $\vec{p}_i$ with the camera's optical axis facing along the normal towards the point. We use the same formulation for defining the \emph{pre-touch} pose of the end-effector, where we recycle the solution for the camera pose. %
    We use~\cite{andersson2019casadi} to obtain precise solutions with minimal deviation and reject any solutions if they deviate by more than $\SI{5}{\milli\meter}$ or $5^\circ$.
    \item Once the IK-solutions were obtained successfully, we move the robot to the first one, take a depth observation $z_i$ and the camera pose $\tf{W}{T}{c,i}$.
    \item Subsequently, we move the robot to its pre-touch pose. We obtain an ambient reading $\hat{\tau}_i$ by averaging $50$ readings collected in this position. We move on to touching the object by moving the end-effector towards $\vec{p}_i$ along $\vec{n}_i$ at $\SI{1.5}{\centi\meter\per\second}$ until we read the relevant deviation of the sensor measurements from the ambient reading.\footnote{Using the deviation from the ambient serves as a consistent calibration.} We collect an averaged reading $\hat{\tau}_i'$ of $40$ samples which we use to calculate the dataset item $\tau_i = \hat{\tau}_i' - \hat{\tau}_i$. Along with the tactile measurement, we also record the pose $\tf{W}{T}{\tau,i} \in SE3$ in which the end-effector made contact with the object.
    \item The patch is extracted by projecting the sensor's corners into the $z_i$ using $\tf{c}{T}{\tau,i} = \tf{W}{T}{c,i}^{-1}\tf{W}{T}{\tau,i}$. We warp the enclosed depth image into a square and resize it to its target resolution, obtaining our raw depth patch $\hat{z}_{d,i}$. 
    \end{enumerate}

\subsection{Objects and Dataset Extension}

We collect data from $8$ primitive objects from a shape-sorting toy. As shown in~\cref{fig:pt_objects}, these objects comprise different geometric features, such as edges of different curvatures and points of different sharpness. We train our model only on data gathered from these objects. We collect $1600$ samples from these objects.
As shown in \cref{fig:setup_and_flow}, we observe a large gap between the real depth patches and their simulated equivalent. This presents a challenge, as our downstream tasks are based on simulated depth data. The real and simulated data need to be very closely aligned. Instead of integrating a more realistic rendering model, we choose to train the model to close the gap between our real and simulated observations. We do so by adding a second sample $(\tau_i, \hat{z}_d)$ for each sample $(\tau_i, z_d)$ in our dataset, with $\hat{z}_d$ being a virtual depth patch obtained from a simulation environment. We normalize both real and simulated dataset by subtracting mean of lowest $10\%$ of depth values from the patch and then normalizing it to a $3mm$ range. We clamp values exceeding the $[0, 1]$ range. We perform this step to exclude depth information which cannot realistically be captured by the sensor.

\section{Experimental Evaluation}
\label{sec:experiments}
We evaluate the utility of our \ourName{} for robotic tasks by using it in  \cref{sec:experiments:object_recognition} to perform object recognition as described in~\cref{sec:technical_approach:object_recognition}. Further, in \cref{sec:experiments:grasp_quality_prediction}, we perform the grasp stability prediction outlined in \cref{sec:technical_approach:grasp_stability_prediction}.

\subsection{Object Recognition}
\label{sec:experiments:object_recognition}

\begin{table}[t]
    \vspace{2mm}
    \setlength{\tabcolsep}{3pt} %
    \centering
    \begin{tabular}{l|cccccccc}
        \toprule
                            && \multicolumn{3}{c}{Primitives Objects}  && \multicolumn{3}{c}{Everyday Objects} \\
        \midrule
                            &&  P & T & P+T && P & T & P+T \\
                            \cmidrule{3-5} \cmidrule{7-9}
    \textit{Human Baseline}  && $34\%$ & $60\%$ & $71\%$ && $51\%$ & $77\%$ & $86\%$ \\
    \midrule
        \ourName{} ($N=5$)   && $12\%$ & $\mathbf{42\%}$ & $\mathbf{42\%}$ && ${60\%}$ & $\mathbf{84\%}$ & ${64\%}$ \\
        \ourName{} ($N=150$) && ${24\%}$ & $\mathbf{47\%}$ & $\mathbf{47\%}$ && $53\%$ & $\mathbf{62\%}$ & $60\%$ \\
        \bottomrule
    \end{tabular}
    \caption{Results of the object recognition experiment. We compare the baselines on two object sets (primitives and everyday) with 3 different input modalities: proprioception (P), touch (T), and both together (P+T). Our proposed \ourName{} is evaluated with ($N=5$) importance sampling as well as without ($N=150$). We also add the human baseline from our user study to put the results in perspective.}
    \label{tab:recognition_results}
    \vspace{-3mm}
\end{table}

For the first downstream task, we evaluate the performance of \ourmodel for the object recognition task. Akin to reaching into a backpack for a pen, the agent needs to identify the correct object out of a given candidate set in this task. 
We assume that the agent has detailed 3D models, such as CAD models or 3D scans of the candidate objects, to predict the sensor measurements that should have been observed when touching an object.
We use the probabilistic inference formulation from \cref{sec:technical_approach:object_recognition} and compare the performance of touch, proprioception (i.e., contact location), and their combination. 
We set the expected variance $\sigma = 8.6$ for touch and $\sigma = 0.0088$ for proprioception. We extract these values from the dataset's validation set. The touch variance is the measured variance on the trained model predictions. In contrast, the proprioception variance is calculated from the collected dataset by measuring the mean distance between the sampled and actual touch location.

{\noindent \textbf{Experimental Setup}:} We perform the experiment on two sets of objects: the primitives from which we also collected our training data and a set of five real everyday objects, shown in \cref{fig:pt_objects}. 
We only use five primitive objects to make the experiments more comparable. 
We pick \emph{square (4)}, \emph{hexagon (6)}, \emph{cylinder (1)}, \emph{star (12)}, and \emph{flower (10)} (numbers reference the number printed on the object).
As we do not focus on pose estimation, we fix the objects to wooden bases so that they do not move nor turn when the robot interacts with them. The 3D scans we take of the objects reflect this setup.
Each experiment involves an object being touched $10$ times in locations sampled according to our approach described in \cref{sec:technical_approach:object_recognition}. We repeat each experiment $5$ times for each object and report the success rate of the three different conditions in correctly identifying the object.
Additionally, we perform an ablation study with all $10\times5\times3\times5=750$ samples we obtain for each object set. We ignore the informed object hypothesis procedure and randomly sample from the collection of all recorded samples according to the ground truth object~\mbox{($N=150$)}. As a result, all modalities draw samples from the same distribution and are thus more comparable.

{\noindent \textbf{User Study}:} Furthermore, we conducted a user study with five participants unfamiliar with this work and with the same object sets to evaluate the performance of the human touch sense on the object recognition task. Each participant is shown a picture of the object sets during the evaluation. We sample an object randomly, let the person touch it blindly 10 times with a single finger, and then record whether they are able to recognize it. We imitate our proprioceptive modality by covering the participant's finger with a stiff piece of tape, to neutralize their sense of touch. For the touch only input modality, we ask the participant to hold their hand still, while the test object is put in contact with their finger from a random direction by the experimenter, obfuscating the spatial relation of contacts. For the combined modality, the participant is allowed to both move their hand and feel the object surface unencumbered. 
We repeat the process for each object set and each modality 7 times, and average the results.

{\noindent \textbf{Results}:} We present the results in \cref{tab:recognition_results}. We find both methods to work well for the everyday object set, especially with the informed object sampling with touch performing $20\%$ better than proprioception. This trend is also present for the primitive recognition task, although at a much lower performance for both methods. 
This is expected, due to these objects' greater similarity in extent and surface features. As we discard the informed object sampling, the gap between the approaches narrows significantly. 
In the case of the primitives, both methods improve, with proprioception benefiting the most. For the everyday objects, the performance of both marginally diminishes. 
This signifies that informed sampling is detrimental to the primitive recognition task, as it can lead to an incorrect hypothesis which it then reinforces due to the similarity of the objects. 
Conversely, in the real object task, it helps due to the dissimilarity of the objects.
Interestingly,  we obtain inferior performance for the combination of both modalities. Upon analyzing the failures, we find that our chosen variance for proprioception is too narrow, leading to an aggressive diminishing of the estimate when the measured end-effector position deviates too much from the prediction. Increasing the variance improves joint prediction, but does not change the relative performance of the methods, which is why we choose not ablate this change in detail here.
Finally, we note how the \ourName{} and the human baseline results follow similar trends, validating our findings. While the human baseline scores on average higher success, the difference is moderate, highlighting the efficacy of our learned system.
In conclusion, we find that our \ourmodel model does transfer to predicting touch signals outside of its original training distribution, achieving $84\%$ accuracy on the tactile object recognition task with everyday household objects.

\subsection{Grasp Quality Prediction}
\label{sec:experiments:grasp_quality_prediction}
For the second downstream task, we show that \ourName{} can be leveraged to generate grasp labels for training a tactile grasp success predictor. We compare such a predictor against a baseline which uses point clouds as input.

{\noindent \textbf{Simulation}:} As described in \cref{sec:technical_approach:grasp_stability_prediction}, we first generate a large-scale dataset using PyBullet as our renderer. We use objects from the ShapeNetSem object set~\cite{savva2015semgeo} and pre-annotated grasps from ACRONYM~\cite{eppner_acronym_2021} which has annotated grasps for more than 8800 objects (the same object instance with a different scale is considered a different object). For each of the objects, we randomly sample five successful and five unsuccessful grasps and record the depth patches $\depthPatch_\mleft, \depthPatch_\mright \in \realSpace^{17 \times 17}$ for each finger using a virtual orthographic camera. 
For each object, we randomly sample ten random camera poses and extract object point clouds. We use the same set of grasps but discard grasps that do not face the camera determined by the angle between the grasp direction and the camera normal. If the angle is above $85^\circ{}$, the grasp is considered invalid. During training, we randomly downsample the point cloud to 2048 points at each batch. 
Overall, due to some imperfect meshes in ShapeNetSem, we have roughly 8500 objects which we split into a train ($70\%$), validation ($10\%$), and test set ($20\%$), giving us a total of 58770 data points for training the tactile network and 253508 data points to train the point cloud network.

\begin{figure}
    \centering
    \includegraphics[height=2.65cm]{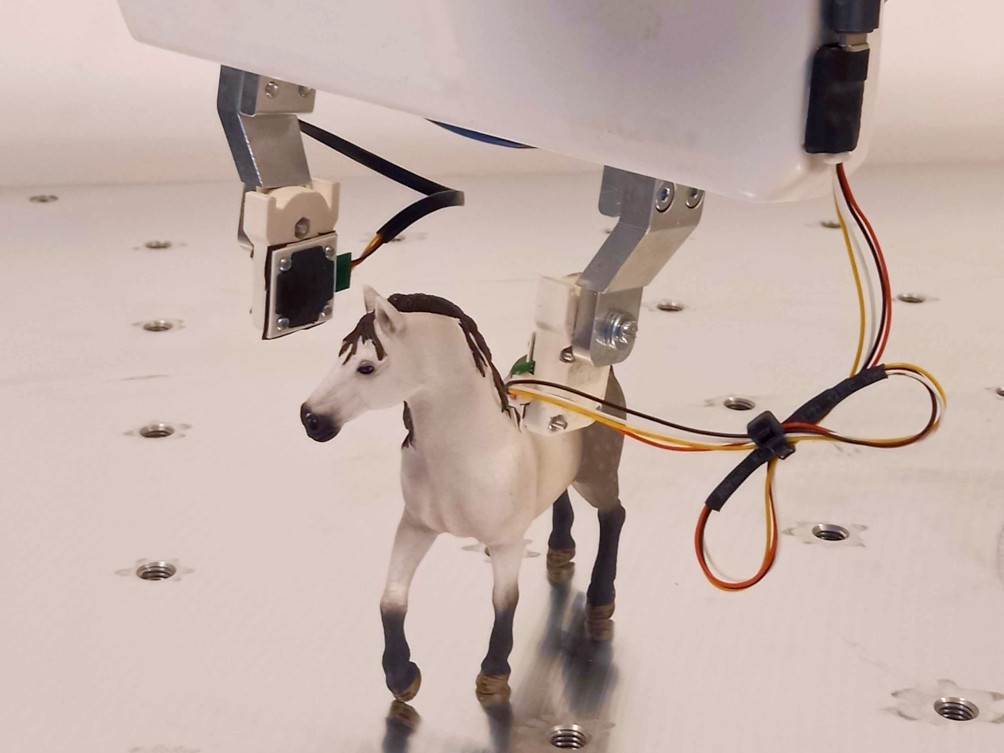}
    \includegraphics[height=2.65cm]{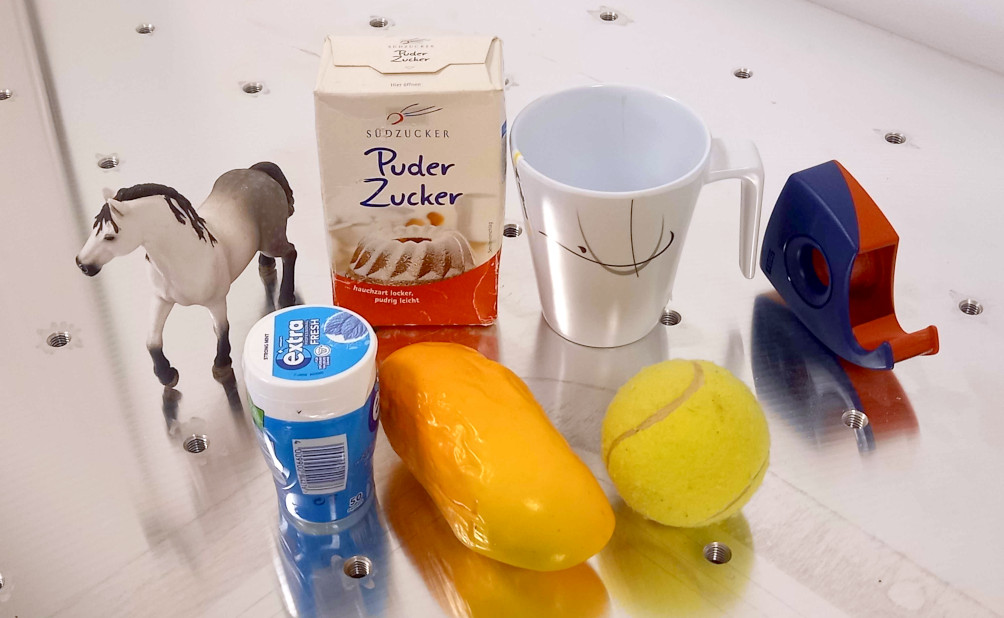}
    \caption{\textbf{Left}: Setup for grasping with \reskin sensors. We 3D-print two fingers for the sensors, which possess suitable bores for mounting the sensors and gel pads to them. The microcontrollers are attached to the gripper, and the wires are left floating loosely so they do not get strained by the movement of the fingers.
    \textbf{Right}: Overview of our objects used in our grasping validation. With the objects, we try to cover a range of different shapes and weights. \emph{Back row}: Toy horse, pack of sugar, cup, tape dispenser. \emph{Front row}: Gum can, mango, tennis ball.}
    \label{fig:reskin_gripper}
    \vspace{-3mm}
\end{figure}

{\noindent \textbf{Real World}:} To demonstrate that \ourName{} easily transfers to the real world, we perform a real grasping experiment (Sim2Real). As shown in \cref{fig:reskin_gripper}, we replace the single ReSkin touch sensor on the robotic gripper with a ReSkin sensor mounted on each finger.
To predict grasps, we use ContactGraspNet~\cite{sundermeyer2021contact}, an off-the-shelf grasping network. We place a single object on a table. After acquiring a suitable grasp, we move the robot towards it and slowly close the gripper until both of the ReSkin sensors measure a significant activation or the gripper is stalled. We then record the signal for both, attempt the grasp, and manually label it as successful or not. Overall, we recorded 76 grasps across 7 objects (refer to \cref{fig:reskin_gripper}).
To extract object point clouds as done in ContactGraspNet, we use \cite{xiang2021learning} to detect the object mask which is then projected into 3D using the depth image and the camera's intrinsic. 

\begin{table}
    \vspace{2mm}
    \centering
    \begin{tabular}{r|cc}
        \toprule
        Modality & Sim Acc. ($\uparrow$) & Real Acc. ($\uparrow$)   \\
        \midrule
        $\graspQualityNetworkPC$ (PC)                & $64\%$ &          $47\%$  \\
        \ourName~(T) + $\graspQualityNetworkPC$ (PC) & $73\%$ &          $36\%$ \\
        \ourName~(T)                                 & {$\mathbf{76\%}$} & {$\mathbf{79\%}$} \\
        \bottomrule
    \end{tabular}
    \caption{Results for grasp quality prediction using different modalities: Point Clouds (PC) from $\graspQualityNetworkPC$ and Tactile (T) from \ourName. For each modality, we report the accuracy of the grasp success prediction.}
    \label{tab:grasp_quality_evaluation}
    \vspace{-3mm}
\end{table}

{\noindent \textbf{Baseline}:}
Our baseline adapts \cite{chumbley_integrating_2022, calandra_feeling_2017} and uses a learned neural network $\graspQualityNetworkPC$ that processes a partial point cloud. The point cloud is first processed with an XYZ-feature-aware PointNet++ backbone~\cite{qi2017pointnet++} before it is down-pooled and concatenated with the grasp $g\in SE(3)$ (represented as $\realSpace^{4 \times 4}$). These features are then given to two fully connected layers to infer the grasp success score $\graspQualityOutput$. We use the same renderer as described above but render point clouds of the object using an external camera, located approximately $70cm$ away.
In addition to $\graspQualityNetworkPC$, we also study the performance of \ourName~(T) + $\graspQualityNetworkPC$ (PC), which is a classifier of the same structure, but with an input space composed of the predicted tactile readings and the point cloud embedding. The aim of this architecture is to study if the two modalities can complement one-another.

{\noindent \textbf{Results}:}
We show the results of our grasp quality predictor in \cref{tab:grasp_quality_evaluation}.
We observe that the model solely trained on touch data outperforms both baseline models. This can be attributed to the fact that the point cloud simulation does not include realistic sensor noise which shows that using a low-dimensional sensor such as the ReSkin transfers more readily to a sim-to-real setup as opposed to higher-dimensional sensors such as cameras. From our results we can also conclude that our tactile-model captures the relevant domain fully, as the combination of tactile and point cloud observations (\ourName~(T) + $\graspQualityNetworkPC$ (PC)) does not outperform the sole tactile model (\ourName~(T)).

\section{Conclusion}
\label{sec:conclusion}

In this work, we investigated using a low-dimensional tactile sensor for general robotic tasks. Unlike their vision-based counterparts, such sensors are attractive for their compact form factor and low cost. However, their low-dimensional readings raise the question of whether they can be used for tasks requiring more advanced tactile distinction.
Towards this goal, we proposed \ourmodel, a novel approach that infers expected tactile readings from small depth images of surfaces. We introduced a low-dimensional model architecture and a procedure for collecting data to train the model, including a data augmentation strategy to improve sample efficiency. By pretraining the model on one set of objects and leveraging it for separate downstream tasks, we demonstrated that our model can transfer successfully outside its object training distribution. We additionally show the potential of Sim2Real applications, reducing time and wear needed to collect real world data.
We view our results as an encouraging step towards using such inexpensive tactile sensors more often in robotics.
For future work, we see an opportunity to build the inverse of our approach: a model predicting depth images from tactile signals to obtain 3D object features. This would enable our approach to perform full tactile 3D reconstruction. We also see a promising direction in integrating \ourName{} together with visual inputs to enable more complex tasks.

\begin{footnotesize}
    \bibliographystyle{ieeetr}
    \bibliography{sources.bib}
\end{footnotesize}

\end{document}